\documentclass[10pt,twocolumn,letterpaper]{article}

\usepackage{cvpr}
\usepackage{times}
\usepackage{epsfig}
\usepackage{graphicx}
\usepackage{amsmath}
\usepackage{amssymb}
\usepackage{multirow}
\usepackage{booktabs}

\usepackage[pagebackref=true,breaklinks=true,letterpaper=true,colorlinks,bookmarks=false]{hyperref}

\cvprfinalcopy % *** Uncomment this line for the final submission

\def\cvprPaperID{017} % *** Enter the CVPR Paper ID here

\ifcvprfinal\fi
\begin{document}

% %%%%%%%%% TITLE
\title{3D Object Detection on Point Clouds using Local Ground-aware and Adaptive Representation of scenes' surface}
\vspace{-0.75 in}
\author{Arun CS Kumar\\
{\tt\small arunkarthikcs@gmail.com}
\and
Disha Ahuja\\
{\tt\small ahuja.disha@gmail.com}
\and
Ashwath Aithal\\
{\tt\small ashwathaithal@gmail.com}
\and
\vspace{-0.2 in}
NIO USA Inc.,
3200 N. First St, San Jose, CA 95134\\
}

\maketitle
\vspace{-0.75 in}
%%%%%%%% ABSTRACT
\begin{abstract}
\vspace{-0.1 in}
A novel, adaptive ground-aware, and cost-effective 3D Object Detection pipeline is proposed. The ground surface representation introduced in this paper, in comparison to its uni-planar counterparts (methods that model the surface of a whole 3D scene using single plane), renders a far more accurate ground representation while being approximately 10x faster. The novelty of the ground representation lies both in the way in which the ground surface of a scene is represented, as well as in the computationally efficient manner in which it is derived. Furthermore, the proposed object detection pipeline builds on the traditional two-stage object detection models by incorporating the ability to dynamically reason the surface of the scene, ultimately achieving a new state-of-the-art 3D object detection performance among the two-stage Lidar Object Detection pipelines.
\end{abstract}
\vspace{-0.2 in}

\begin{figure*}
\vspace{-0.15 in}
\begin{center}
\center{\includegraphics[width=\textwidth]{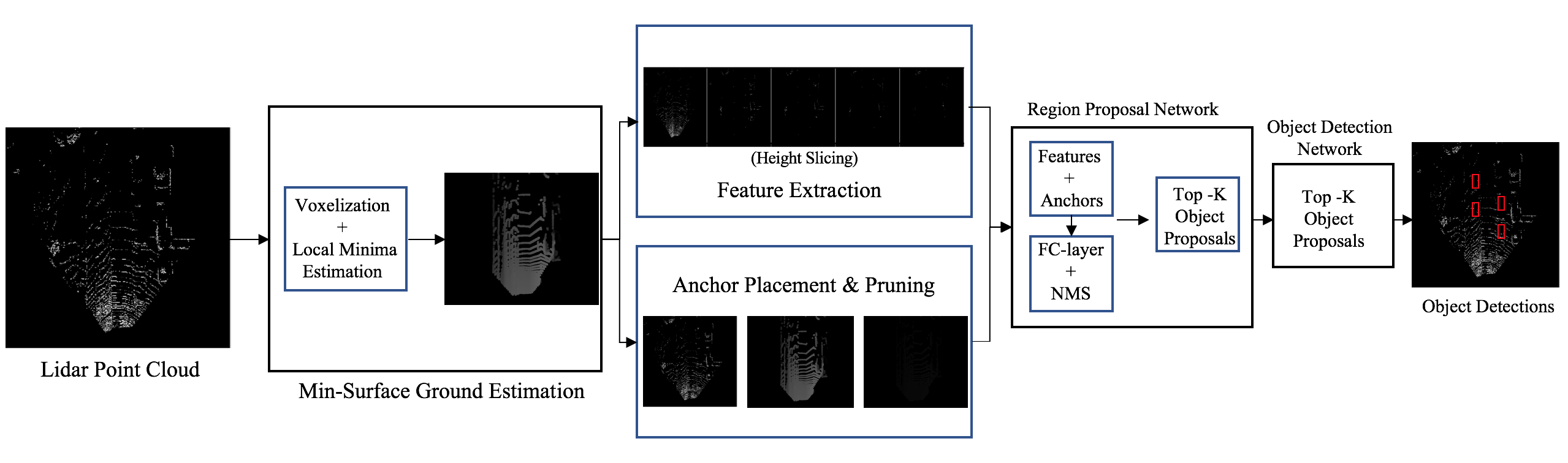}}
\end{center}
   \caption{Pipeline of the proposed architecture: Given a lidar point cloud, the ground surface-based ground estimation is computed. Using the ground estimate, feature extraction and anchor generation is performed, which are then fed to the RPN and OD networks, which outputs Object Detection.}
   \vspace{-0.1 in}
\label{fig:pipeline}
\end{figure*}

%%%%%%%%% BODY TEXT
\section{Introduction}\label{sec:introduction}
 \vspace{-0.05 in}
Building robust scene understanding pipelines is one of the most critical components of robot navigation. In that respect, one of the central problems that the Autonomous Navigation community is faced with, is the accurate detection and localization of objects in 3D. Among several modalities of data that are being employed, Lidar stands out to be one of the most robust ones, with rich range information readily available.  
Thus, in recent times, the advancements in Lidar technology have pushed the computer vision community to build perception algorithms that work purely on Lidar point clouds. In that regard, the goal of this paper is to build a cost effective pipeline, that is capable of reasoning the scene more adaptively, in order to perform 3D object detection purely using sparse Lidar point clouds, without any reliance on RGB/camera data. 

Using images in isolation or image and lidars~\cite{chen2017multi, ku2018joint} in a combined framework has a number of drawbacks when employed in practice: Camera images are rendered unreliable when visibility is poor such as during {\it fog} or {\it rain} or at {\it night}. On the other hand, lidars are shown to be far more effective and reliable under such extreme weather conditions.
In addition, existing object detection pipelines that are learned as a multi-modal networks especially with early fusion such as~\cite{ku2018joint}, while tend to perform better than architectures where lidars or images are used in isolation, but curtail the redundancy, which is critical for the real-time functioning of most autonomous systems.

In this paper, we propose an end-to-end trainable 3D object detection pipeline that uses piece-wise fit ground estimates that is both accurate and computationally inexpensive. Ground points segmentation is a vital component of a number of Lidar-based perception pipelines, such as 3D Object Detection, Drivable/Free Space Estimation, Occupancy Grid calculation, to cite a few. The proposed ground representation is generic and can be directly employed for ground estimation for sparse point clouds and can also be used directly in any 3D Object Detection pipeline. 

We propose a novel and computationally efficient approach that computes several local piece-wise ground representations of the scenes’ surface making the representation more accurate, while in comparison, being ~$\sim$10x faster than its uniplanar counterparts (Section ~\ref{ssec:implementation_details}). The novelty of the proposed ground representation lies both in the way in which the ground plane of the scene is represented in Lidar perception problems, as well as in the (computationally efficient) way in which we derive the representation. Unlike most geometry-driven ground point segmentation algorithms ~\cite{harakeh2015ground, chen20153d}, the proposed ground surface estimation method (referred to as ground surface) has no dependency on the nature of lidar or the lidar-scene interaction, thus need no prior knowledge about the lidar position or its type, and can work equally efficiently for all types of lidars (rotating or solid-state). Furthermore, the computational cost for most ground segmentation methods~\cite{chen2017multi, bogoslavskyi2017efficient} linearly increase as the number of Lidars increase, or the lidar point clouds’ density increase, whereas the proposed method works nearly on a constant time irrespective on the type or the number of lidars used.

We extend conventional {\it two-stage} object detection frameworks such as AVOD~\cite{ku2018joint}, by incorporating the ability to reason scene's surface more dynamically, ultimately building an end-to-end trainable architecture, that is {\it 25\%} faster, and achieves state-of-the-art performance in 3D object detection among two-staged Lidar object detectors.

\section{Related Work}\label{sec:related_work}
 \vspace{-0.05 in}
\subsection{3D Object Detection}\label{ssec:object_detection}
 \vspace{-0.05 in}
Most of the literature for the problem of Object Detection on point clouds can be categorized into two major subsets: (1) Single Shot detectors that directly operate on 3D point clouds~\cite{qi2017pointnet,shi2019pointrcnn,zhou2018voxelnet} and (2) Two-stage object detection models that often downsizes the 3D data to a 2D representation~\cite{ku2018joint, chen2017multi}. While methods that employ 2D downsizing of the 3D point clouds are relatively less accurate in comparison to their 3D counterparts, but are far more efficient to be used in real-time applications. 

Single shot detectors are end-to-end trained pipelines that directly operate on raw point clouds and compute object detection or point classification~\cite{qi2017pointnet,shi2019pointrcnn,zhou2018voxelnet, li20173d}. VoxelNet~\cite{zhou2018voxelnet}, one of the recent works, divides point clouds into equally spaced voxels, and computes a transformed feature representation termed as Voxel Feature Encoding (VFE). However, the major overhead with~\cite{zhou2018voxelnet} is that the initial layers employ 3D convolutional layers, which are computationally expensive. More recently, PointNet~\cite{qi2017pointnet} computes point classification by consuming unordered Lidar points and regressing point-wise classification. PointRCNN~\cite{shi2019pointrcnn} builds on top of ~\cite{qi2017pointnet} extending PointNet to a {\it 2-staged} object detection pipeline in order to be employed on real world Lidar data. Since these methods can directly consume unordered point clouds, there is no need for voxelization or 2D/3D grid representation. However, the overall inference times of these methods~\cite{qi2017pointnet,shi2019pointrcnn} are considerably high in comparison to the proposed 2-stage object detector. 

On the other hand, {\it two-stage} detectors are typically Faster-rcnn~\cite{ren2015faster} based models such as~\cite{ku2018joint, chen2017multi}. AVOD~\cite{ku2018joint}, one of the commonly used architectures, that leverages multimodal training and inference framework, where both image and Lidar data are used to perform 3D object detection. The first stage comprises the Region Proposals generation layer (referred to as Region Proposal Network (RPN)) that quantifies the objectness in the scene, which is subsequently followed by a Object Detection layer that uses the candidate object proposal provided by the RPN layer to regress the object bounding boxes.

Overall, in comparison, though single shot detectors~\cite{qi2017pointnet, shi2019pointrcnn, zhou2018voxelnet, li20173d, liu2016ssd} have a slight edge over their two-staged counterparts, when it comes object detection accuracy, the two-stage object detection algorithms~\cite{ku2018joint} often tend to be far more efficient (computationally) with a comparable performance. 
There have been advances since~\cite{ku2018joint} in bridging the gap between 3D point cloud and 2D feature representations. One of the works is~\cite{lang2019pointpillars} that downsizes the 3D point clouds to 2D representations by converting it to a 2D point pillars (or 2D grids of features), features of 3D points, with known length 2D vectors placed vertically on the {\it xz}-plane. This enables the use of 2D convolution in place of 3D, thus making it relatively more efficient than~\cite{zhou2018voxelnet}. 

In terms of feature representation, the proposed method is very relevant to~\cite{ku2018joint}, uses height slices as a means to render 2D voxelization of 3D features for efficiently downsizing the 3D features to 2D, in a two-staged architecture. The primary difference between the proposed model and~\cite{ku2018joint} stems from the way in which the surface of the scene is modeled. In addition, feature representation such as 2D voxelization are quite generic and thus are easily transferable and can function across different types of data (point clouds from different sensors) as shown in Figure~\ref{fig:od_diff}. For example, a model that is trained on a single velodyne sensor setup works seamlessly or point clouds procured from multiple velodyne sensors as shown in Figure~\ref{fig:od_diff}.
 \vspace{-0.05 in}

\subsection{Ground Segmentation}\label{ssec:ground_segmentation}
 \vspace{-0.05 in}
In the context of Object Detection, most existing frameworks especially the 2-stage pipelines such as FasterRCNN~\cite{ren2015faster}, rely heavily on accurate ground representation for robust feature extraction (for both deep-learned or height slicing-based) and/or for placing anchor boxes accurately on ground. Oftentimes, the estimation of ground is achieved via segmenting ground points ~\cite{harakeh2015ground, chen20153d, bogoslavskyi2017efficient, zhou2018voxelnet, velas2018cnn}, and is typically followed by fitting a 3D plane to the scene (mostly using RANSAC)~\cite{chen20153d}. But this method has a few major drawbacks: 
\vspace{-0.075 in}
\begin{itemize}
    \item A real-world scene is often too sophisticated to be represented using a single plane, as such overly simplistic assumptions about the scene ultimately leads to deterioration in the overall 3D object detection performance.
    \vspace{-0.075 in}
    \item The cost of fitting a plane to the scene using conventional methods is significantly high.
    \vspace{-0.075 in}
    \item Even when not used a planar representation, the cost of segmenting ground points by itself, even with most state-of-the-art methods, is significantly high.
    \vspace{-0.05 in}
\end{itemize}

A variety of approaches have been attempted ranging from complex geometric reasoning~\cite{chen20153d, bogoslavskyi2017efficient} (reasoning conditioned on Lidar-scene interaction), to stereo camera-driven approach to compute probabilistic fields~\cite{harakeh2015ground}, to data-driven approaches to train expensive deep learned models ~\cite{velas2018cnn, zhang2018efficient}. 

In this paper we propose,
\vspace{-0.1 in}
\begin{itemize}
    \item An adaptive and piece-wise fit ground surface representation that is {\it $\sim$10x} faster and considerably more accurate for ground segmentation (Section \ref{sec:evaluation}).
    \vspace{-0.075 in}
    \item An end-to-end trainable Object Detection pipeline, that builds on the ground surface representation, yielding state-of-the-art performance among the similar class (two-staged object detection) of object detection architectures.
    \vspace{-0.075 in}
\end{itemize}

\section{Model}\label{sec:model}
\vspace{-0.05 in}
Given a set of lidar points, the {\textit Ground Segmentation} (Section~\ref{ssec:ground_segmentation}) technique identifies the non-ground points from the point cloud, which is followed by {\it 3D Object Proposal generation} (Section~\ref{ssec:3d_region_proposals}) that leverages the ground estimate obtained via Ground Segmentation Algorithm (Section~\ref{ssec:ground_segmentation}) for generating and placing candidate 3D object proposals on the estimated ground. The {\it 3D region proposals} is fed to the {\it Region proposals} (Section~\ref{ssec:rpn}) along with the features extracted using (Section~\ref{ssec:od_layer}). Subsequently, the output of the {\it Region Proposals Network} are then fed to the {\it Object Detection Network} (Section~\ref{ssec:od_layer}), that provides the object bounding boxes.

\subsection{Adaptive Ground Segmentation}\label{ssec:adaptive_ground_segmentation}
\vspace{-0.05 in}
In this section, we propose a general purpose ground point segmentation algorithm for sparse lidar point clouds, that computes a piece-wise local ground representation in a computationally efficient manner, as opposed to using planar ground representation (single plane to represent the scene). In contrast to the traditional planar representation of the ground~\cite{ku2018joint}, the proposed method aims to identify the local minima for each 3D point cloud.

Given a 3D lidar point cloud, the proposed method bins the {\it xy}-points to form 2D grid and stores the maximum {\it z} (height) in a 2D matrix denoted as {\it height maps} (as there can be multiple {\it z} values for each {\it xy}, among which we store only the maximum). 

\begin{figure}
{\includegraphics[width=\linewidth]{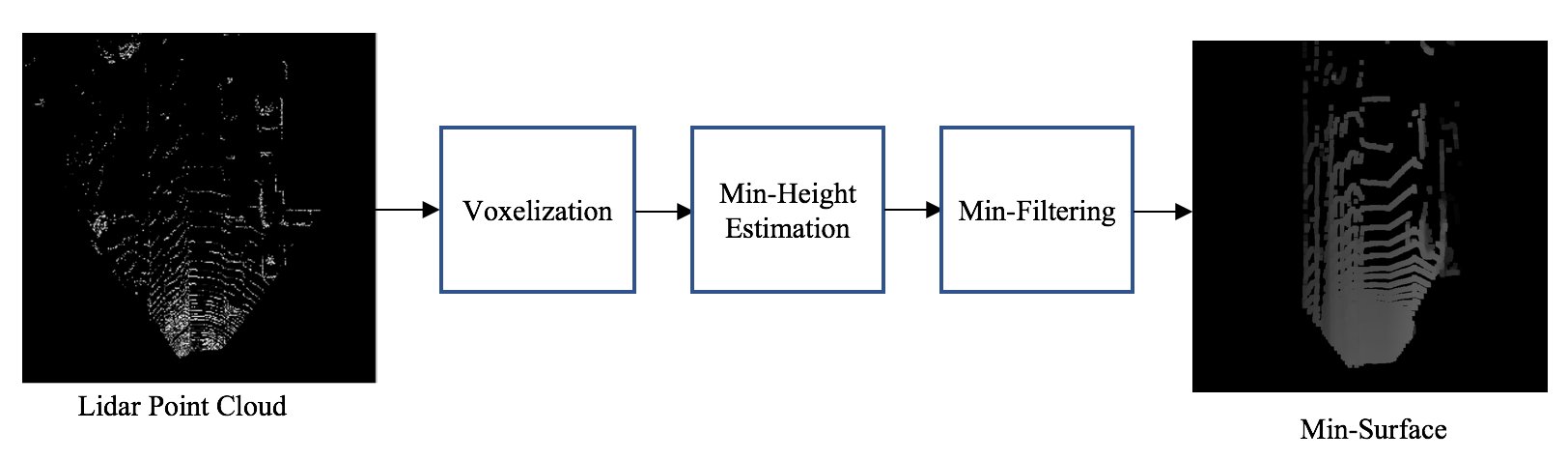}}
   \caption{Pipeline of the proposed ground surface estimation algorithm. Given the lidar point clouds, we first compute voxel grid that assigns each 3D point to a 2D {\it xy} coordinate. It is then followed by height map computation and min-filtering to render the ground surface estimate.}
\label{fig:min_surface_pipeline}
\end{figure}

Then, for each valid {\it xy}-point (non-empty point in the 2D Grid), we search the local neighbourhood to find the minimum height (in a rectangular region), as shown in Figure~\ref{fig:min_surface_pipeline}. Owing to the sparsity of the lidar points and the nature of reflectance of the lidar point cloud, the local neighbour search almost always returns a ground point or at least a point closest to the ground.

Computing maximum of the height map assures that the maximum {\it x}-value of a non-ground region/object may be the top (or non-ground portion) of the region/object, whereas the highest {\it z}-value of the ground is still the ground. This idea is exploited in order to search the neighborhood to identify the ground point estimate of each 3D point. Doing so, allows us to pick the ground point reliably for every local region, when used min-filtering in local neighbourhood. The proposed method is non-parametric (except for the filter size), and involves a single convolution operation, and thus is significantly faster than the geometry based~\cite{chen20153d} methods for ground segmentation.

\subsection{3D Region Proposals}\label{ssec:3d_region_proposals}
\vspace{-0.05 in}
Like most existing {\it 2-Stage} Object Detection algorithms~\cite{ku2018joint}, the proposed method consists of an object candidate generation layer that generates 3D object bounding boxes, also referred to as Anchor boxes ~\cite{ren2015faster}. An anchor box is simply a vector that encodes the bounding box information, the center of the bounding box (in 3D), aspect ratio, and its orientation. There are various ways of representing a bounding box, some of the common ones are 3D Box representation {\it \{x, y, z, l, w, h, $\theta$\}} where {\it \{x, y, z\}} is the center of the bounding box and {\it \{l, w, h\}} are {\it length, width} and {\it height} of the bounding box, and $\theta$ denotes the orientation along the {\it y}-axis. An alternative representation (used in~\cite{ku2018joint}) is 4CA (4 corners and angle) denoted using {\it \{x1, y1, x2, y2, h1, h2\}} where {\it \{x1, y1, x2, y2\}} are top-left and bottom-right corners of the bounding box whereas {\it \{h1, h2\}} are the top and bottom of the bounding boxes. The choice of representation is determined primarily based on the choice of loss function as detailed in Section~\ref{ssec:training}. 

\begin{figure*}
\vspace{-0.1 in}
\begin{center}
\center{\includegraphics[width=\textwidth, height=6cm]{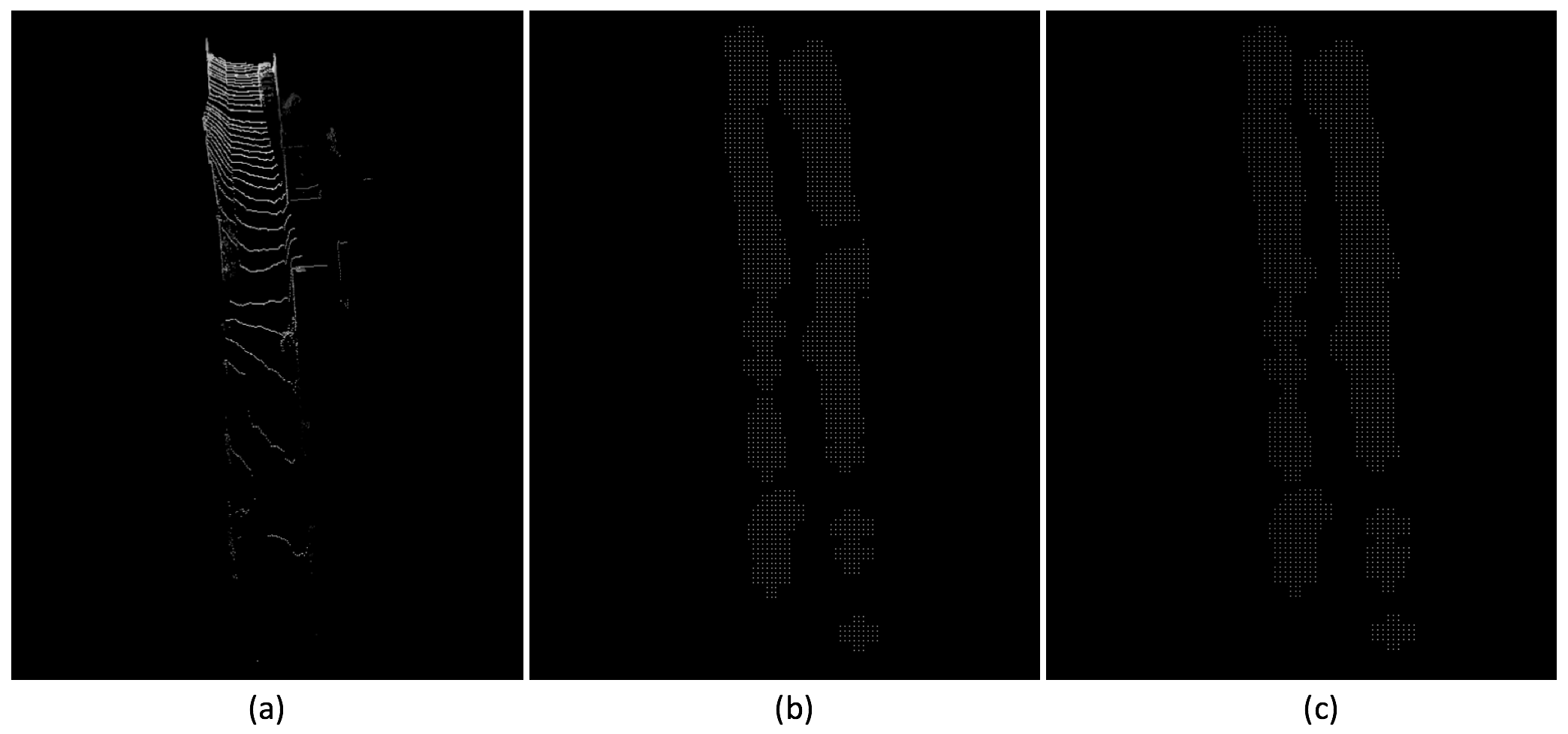}}
\end{center}
   \caption{Comparison of anchor placement for a BEV frame (a) and pruning using ~\cite{ku2018joint}(b) and the proposed method (c).}
\label{fig:anchor_pruning}
\end{figure*}

Unlike~\cite{ren2015faster} that places anchor boxes of different aspect ratio sizes and orientations across the 2D image grid, in order to perform 3D Object detection the region proposals has to be extended to 3D space. However, directly extending the method to 3D results in a significant increase in the number of bounding boxes, which is certainly not scalable. Mv3D~\cite{chen2017multi} and AVOD~\cite{ku2018joint} propose to place anchors more efficiently by estimating the ground or surface of the scene represented as 3D plane using planar coefficients, and the anchor boxes are placed only on the 3D plane. It is evident that the objects of interest lie on the ground, and thus other regions can be largely discarded, ultimately leading to a significantly reduced number of anchor boxes and thus a largely improved inference time.

In the proposed approach, instead of using a uniplanar representation of the scene surface, we adopt a more continuous and piece-wise fit ground representation, to place anchors. Anchor placement is done by sampling the 2D Birds Eye View (BEV) grid evenly, where we used a sample size of {\it 0.5} metres across both {\it x} and {\it y} axis. The anchor placement is followed by Anchor pruning, which largely involves removing anchors that do not have any 3D lidar point in the corresponding voxel, further downsizing the number of anchors used at inference. While it makes sense to simply remove any anchor point centered in a voxel that has no 3D lidar point, it may also lead to missing objects that may be trapped in between two lidar beams altogether. AVOD~\cite{ku2018joint} handles this efficiently by using integral images, which counts the number of 3D points within a bounding box placed at every {\it xy} position of the 2D grid, and the box is considered a candidate if the total number of 3D points that lie within the region is greater than a predetermined threshold. However, using integral image based method requires a valid {\it z}-value for every {\it (x,y)}-point in the BEV. When used a planar representation of the ground as used by Mv3D~\cite{chen2017multi} and AVOD~\cite{ku2018joint}, it is feasible to compute a {\it z}-value for every {\it (x,y)} coordinate, which is not possible using the surface representation proposed in our method. To counter that, we simply use a parameter-free morphological filtering based interpolation to extend ground estimate ({\it z}-value) to K-nearest voxel coordinates, where $k=d/2$ is simply the half length of the diagonal $d$ of the bounding box. The comparison between the points estimated using the proposed method is shown in Figure~\ref{fig:anchor_pruning}.

\begin{figure*}
\vspace{-0.1 in}
\begin{center}
\center{\includegraphics[width=\textwidth]{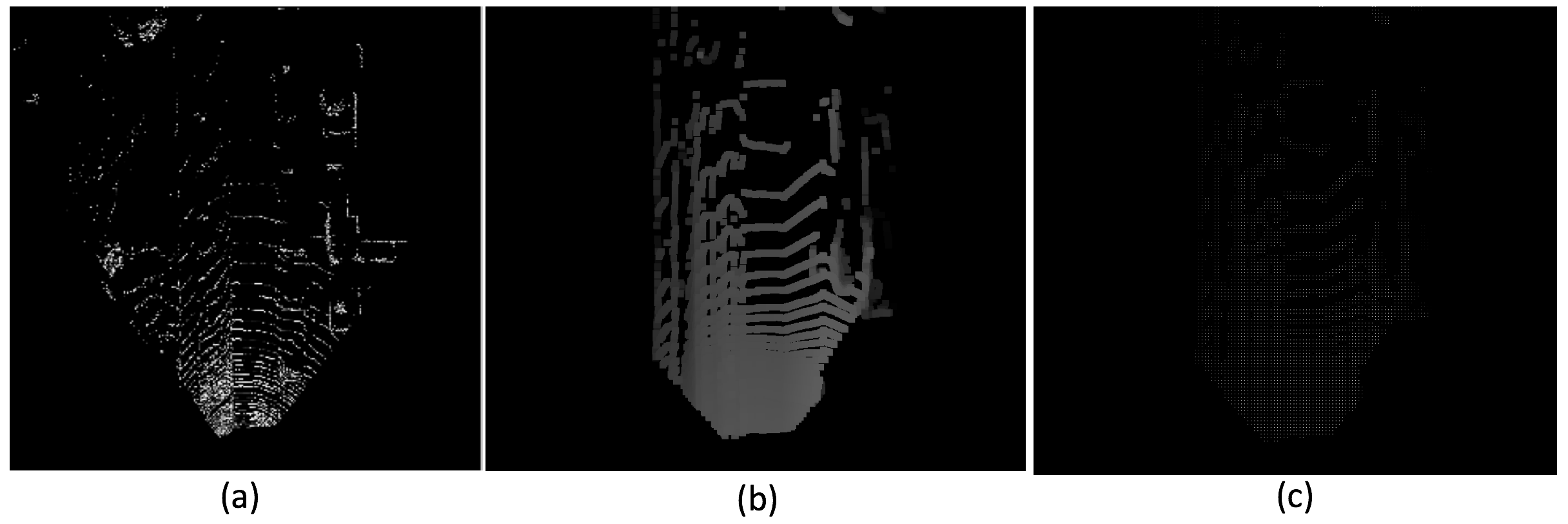}}
\end{center}
   \caption{Given the lidar point cloud {\it (a)}, the ground surface estimated using the proposed ground estimation algorithm is shown in {\it (b)}, and the corresponding anchor generation is shown in {\it (c)}. The shade in {\it (b)} represents the different heights of the ground. {\it (c)} demonstrates the effect of anchor pruning in terms of reduction in the number of anchors.}
\label{fig:short}
\end{figure*}

\subsection{Feature Extraction}\label{ssec:feature_extraction}
\vspace{-0.05 in}
Given a lidar point cloud, we utilize the 2D BEV grid computed in Section~\ref{ssec:ground_segmentation} and the estimated ground plane, to compute a downsized 2D representation. As mentioned earlier, using 3D points directly forces the convolutional layers to use 3D convolutions, which is indeed computationally expensive. Thus, we adapt the feature representation used by~\cite{ku2018joint} and represent the 3D point cloud as coarsely voxelized BEV-slices. The primary difference between the height slicing used by~\cite{ku2018joint} and ours is that, we perform height slicing using the finer ground surface estimate as opposed to the ground plane used by~\cite{ku2018joint}. 

% \subsection{Region Proposal Network}\label{ssec:region_proposal_network}

% \subsection{Object Detection Network}\label{ssec:object_detection_network}

\subsection{Network Architecture}
\vspace{-0.05 in}
\subsubsection{Region Proposal Network}\label{ssec:rpn}
\vspace{-0.05 in}
Following the feature extraction and ground segmentation steps, the features are then fed into the Region Proposal Network and the outputs of the Region Proposal Network is then subsequently fed into the Object Detection Layer. 

The Region proposal stage comprises of two components, the first stage crops and resizes the final convolutional layer features corresponding to each anchor that results in a feature of size {\it 7 x 7 x 32}. The features are then reshaped into a vector and are fed to fully connected layers. The output of the fully connected layers is followed by Non-Maxima Supression (NMS) step, that removes overlapping and redundant bounding boxes. After the NMS step, we pick the top-$N$ anchor boxes for the object detection (In our experiments we used N=300). 

\subsubsection{Object Detection Network}\label{ssec:od_layer}
\vspace{-0.05 in}
The features of the top-$N$ bounding boxes are then fed into the Object Detection Network, which outputs the {\it \{x, y, z, l, w, h, $\theta$\}}. Based on candidate 3D proposals, the Object detection Network learns to predict the bounding box coordinates, size and orientation. The outputs are again fed into a NMS stage that prunes overlapping bounding boxes, which then outputs the object detections.

\vspace{-0.05 in}
\section{Dataset}\label{sec:dataset}
\vspace{-0.05 in}
We used KITTI Raw and Object detection datasets~\cite{Geiger2012CVPR}, to train and evaluate our model. The KITTI Object detection datasets is provided with 3D Lidar object annotations by~\cite{Geiger2012CVPR}. However, the annotations are available only for camera view. In other words, the annotations are available only for the front view of the ego-vehicle for which the RGB data is procured. In order to train a network to compute object detection on a complete 360$^{\circ}$, we annotated all Lidar frames of KITTI Raw dataset with 3D bounding boxes and it corresponding orientation. The format of the annotations we performed on KITTI Raw is the same as that of the KITTI object detection dataset, and thus can be seamlessly used in training pipelines. The KITTI Raw dataset consist of a total of {\it 43,628} frames, of which {\it 33,292} were used for training and validation and {\it 10,336} are used for testing. The 3D object annotations are performed for all {\it 43,628} frames, for a range spanning {\it 75m}  $\times$ {\it 50m} alongside {\it y} \& {\it x} directions respectively (in BEV), around the ego vehicle, totalling {\it 150m}$\times${\it 100m} as the overall range for annotations across {\it yx}-plane.

\begin{figure*}
\begin{center}
\center{\includegraphics[width=\textwidth]{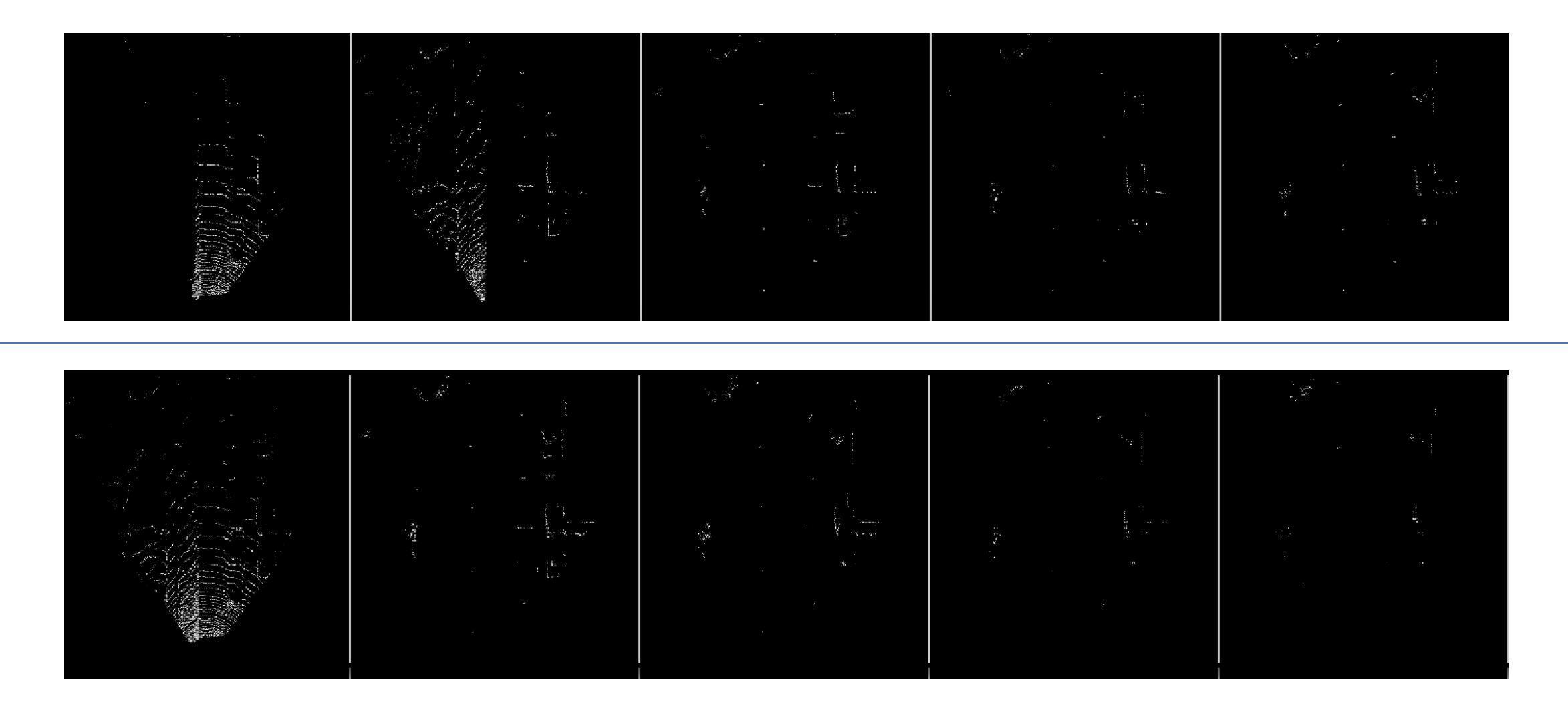}}
\end{center}
   \caption{Comparison of height features using ~\cite{ku2018joint}(top) vs. the proposed method (bottom). It is evident that the features obtained via proposed method (below) results in more consistent and desirable height slicing.}
\label{fig:height_features}
\end{figure*}

Following that, we also categorized the objects annotated as {\it easy, medium} and {\it hard} based on two primary criteria; the object's distance from the Lidar, and the 3D Lidar point density of the annotated object. The categorization of the level of difficulty is slightly different from that of the default KITTI object detection datasets', which bins objects to its level of difficulty based on occlusion information available via RGB camera. Unfortunately, RGB information is available only for the front view of the scenes in KITTI, thus our annotations are done on Lidar point cloud alone, with whose levels of difficulty are categorized in an aforementioned manner. Also, such categorization is primarily designed to ensure the performance of Lidar only object detection pipelines such as the proposed method that do not rely on RGB data.

\subsection{Data augmentation}\label{ssec:data_augmentation}
In addition, we use several augmentation techniques ranging from flipping the data, to generating rotated version of the frames. We also created additional annotations by sampling a few ground truth data and placing them on free spaces in relatively empty frames (object-less regions). For that purpose, we use ground estimation algorithm, to identify ground points that are object-free using the known object annotations, to subsequently replace the ground points with randomly sampled object points. In other words, we identify frames with less objects and place objects (transfer all object points in 3D) on empty regions, creating additional training data.

\begin{table*}[tbp]

\caption{ Comparison of the proposed method against MV3D~\cite{chen2017multi},VoxelNet~\cite{zhou2018voxelnet}, F-PointNet~\cite{qi2018frustum} and AVOD~\cite{ku2018joint}.}
 \label{tab:Compare} 
\centering
    \begin{tabular}{ c | c | c | c | c | c | c | c }
    \toprule
    \multirow{2}{*}{Method} 
      & \multicolumn{3}{c|}{$AP_{3D}(\%)$} 
          & \multicolumn{3}{|c|}{$AP_{BEV}(\%)$} \\ 
          
    \  &  Easy & Moderate & Hard & Easy & Moderate & Hard & Runtime (s) \\ \midrule    
    {\it MV3D~\cite{chen2017multi} } & 71.09 & 62.35 & 55.12 &  86.02 & 76.90 & 68.49 & 0.36\\ 
    {\it VoxelNet~\cite{zhou2018voxelnet} } & 77.47 & 65.11 & 57.73 &  89.35 & 79.26 & 77.39 & 0.23 \\ 
    {\it F-PointNet~\cite{qi2018frustum} } & 81.20 & 70.39 & 62.19 & 88.70 & 84.00 & 75.33 & 0.17\\ 
    {\it AVOD~\cite{ku2018joint} } & 73.59 & 65.78 & 58.38 & 86.80  & 85.44 & 77.73 & 0.08\\ 
    %{\it AVOD (Feature Pyramid)} & 81.94  & 71.88 & 66.38 & 88.53 & 83.79 & 77.90\\
    {\it Ours* } & 70.79 & 60.63 & 55.49 & 87.32 & 78.80 & 78.05 & 0.06\\ 
   
    \bottomrule
    \end{tabular} 
    \vspace{-0.2cm}
\end{table*}

\begin{table}[tbp]
\caption{ Comparison of the proposed method against AVOD~\cite{ku2018joint} on BEV and 3D object detection tasks, on KITTI Raw dataset~\cite{Geiger2013IJRR}. The categorization of {\it easy} and {\it hard} are performed based on the distance of the object from the Lidar center. Objects that are $\leq${\it 20m} from the Lidar center are categorized as {\it easy} and the ones that lie between {\it 20m} and {\it 50m} are categorized as {\it hard}. (note: *-denotes use of Lidar only pipeline due to unavailability of RGB data for 360$^{\circ}$ scene.}
 \label{tab:compare_distance} 
    \begin{tabular}{ c | c | c | c | c }
    \toprule
    \multirow{2}{*}{Method} 
      & \multicolumn{2}{c|}{AVOD~\cite{ku2018joint}* } 
          & \multicolumn{2}{c}{Ours*} \\
    \  &  Easy & Hard & Easy & Hard  \\ \midrule    
    {\it BEV Object Detection } & 89.26 & 59.46 & 89.87 & 78.35 \\ 
    {\it BEV Heading } & 89.60 & 57.93 & 89.69 & 77.60\\ 
    {\it 3D Object Detection } & 76.83 & 37.37 & 76.01 & 53.13 \\ 
    {\it 3D Heading } & 76.67 & 38.84 & 75.89 & 52.80 \\ 

    \bottomrule
    \end{tabular} 
    \vspace{-0.2cm}
\end{table}

\begin{figure*}
\begin{center}
\center{\includegraphics[width=\textwidth]{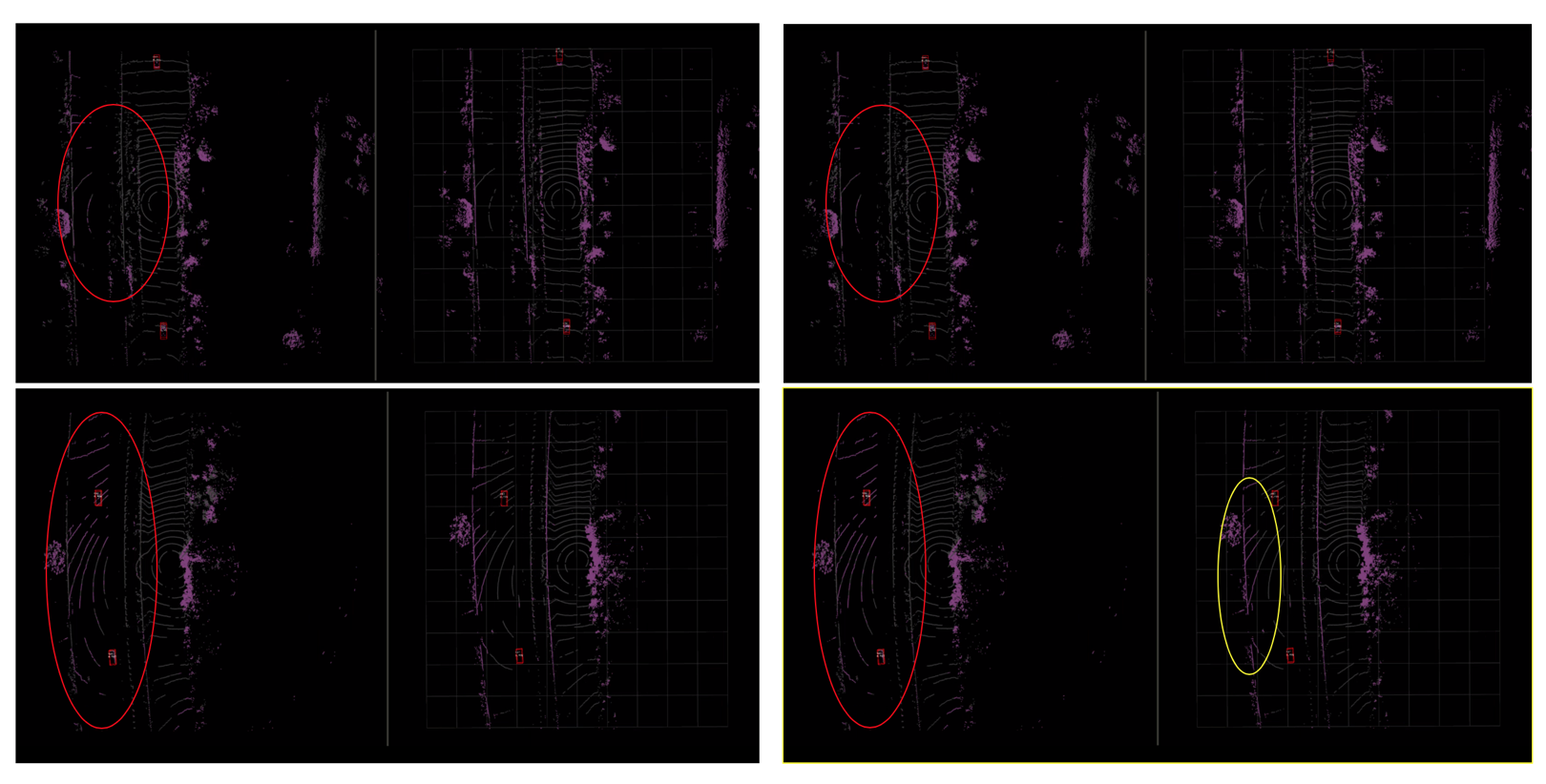}}
\end{center}
   \caption{Qualitative representation of the ground point segmentation comparing planar representations~\cite{harakeh2015ground} {\it vs.} proposed ground surface representation. The highlighted (red) region (left) shows points that are mis-classified as ground due to planar representation used in~\cite{ku2018joint}. ({\it note:} In bottom-right image, the classification range in the {\it x}-axis is only {\it 30m}, thus the ground points that appear to have been mis-classified (highlighted in yellow) are simply points that lie outside the region of interest).}
\label{fig:min_surface_perf}
\end{figure*}

\vspace{-0.05 in}
\section{Training}\label{ssec:training}
\vspace{-0.05 in}
We train and evaluate the proposed architecture using both KITTI Raw and object datasets~\cite{Geiger2012CVPR, Geiger2013IJRR}, for camera view and 360$^{\circ}$. 
The KITTI object dataset consists of {\it 7481} training images and {\it 7518} testing images, and KITTI Raw dataset consists of {\it 33,292} training and validation and {\it 10,336} testing images. 

As mentioned above, we have a {\it 2-staged} training process, where the first stage acts as a Region Proposal generator, that consumes the Feature Maps (Section~\ref{ssec:feature_extraction}) and Region Proposals (Section~\ref{ssec:rpn}), and learns to choose top-N candidate object proposals (in our experiments N is set to 300). The Region Proposal Network is then followed by an Object Detection Network, that learns to regress the center, aspect ratio and orientation of the 3D object bounding box. Both Region Proposal Layer and Object Detection layers are followed by Non-Maxima Suppression that prunes redundant bounding boxes.  

\begin{table*}[tbp]
\caption{ Comparison of the proposed method against AVOD~\cite{ku2018joint} on BEV and 3D object detection tasks, on KITTI Raw dataset~\cite{Geiger2013IJRR}. The categorization of {\it easy, moderate} and {\it hard} are performed based on the density of the Lidar points within the ground truth bounding box. Ground truth object bounding boxes with {\it $\geq$685} points are considered easy, and {\it $\geq$126 pts \& $\leq$685 pts} \& {\it $\leq$126 pts} are categorized as {\it moderate} and {\it hard} respectively. (note: *-denotes use of Lidar only pipeline due to unavailability of RGB data for 360$^{\circ}$ scene.}
 \label{tab:compare_density} 
 \centering
    \begin{tabular}{ c | c | c | c | c | c | c }
    \toprule
    \multirow{2}{*}{Method} 
      & \multicolumn{3}{c|}{AVOD~\cite{ku2018joint}* } 
          & \multicolumn{3}{c}{Ours*} \\
    \  &  Easy & Moderate & Hard & Easy & Moderate & Hard  \\ \midrule    
    {\it BEV Object Detection } & 90.04 & 87.29 & 40.63 & 96.40 & 89.11 & 66.48 \\ 
    {\it BEV Heading } & 89.95 & 86.43 & 39.21 & 96.32 & 88.76 & 65.30\\ 
    {\it 3D Object Detection } & 82.36 & 69.58 & 20.51 & 80.78 & 71.50 & 36.57 \\ 
    {\it 3D Heading } & 82.27 & 69.04 & 19.00 & 80.75 & 71.27 & 36.11 \\ 
    \bottomrule
    \end{tabular} 
    \vspace{-0.2cm}
\end{table*}

\begin{table*}[tbp]
\caption{ Comparison of the proposed method against AVOD~\cite{ku2018joint} on BEV and 3D object detection tasks, on KITTI Raw dataset~\cite{Geiger2013IJRR}, where the IOU threshold for a valid object is set to {\it 0.5} instead of {\it 0.7} used in Table~\ref{tab:compare_density}. The categorization of {\it easy, moderate} and {\it hard} are performed based on the density of the Lidar points within the ground truth bounding box. (note: *-denotes use of Lidar only pipeline due to unavailability of RGB data for 360$^{\circ}$ scene.}.
 \label{tab:compare_density_05iou} 
 \centering
    \begin{tabular}{ c | c | c | c | c | c | c }
    \toprule
    \multirow{2}{*}{Method} 
      & \multicolumn{3}{c|}{AVOD~\cite{ku2018joint}* }
          & \multicolumn{3}{c}{Ours*} \\
    \  &  Easy & Moderate & Hard & Easy & Moderate & Hard  \\ \midrule    
    {\it BEV Object Detection } & 97.44 & 88.22 & 68.26  & 98.25 & 90.55 & 80.09 \\ 
    {\it BEV Heading } & 89.95 & 96.95 & 86.20 &  98.17 & 90.23 & 78.20\\ 
    {\it 3D Object Detection } & 86.03 & 73.07 & 46.21  & 90.24  & 89.73 & 76.66 \\ 
    {\it 3D Heading } & 85.63 & 71.28 & 43.34  & 90.19 & 89.39 & 74.94 \\ 
    \bottomrule
    \end{tabular} 
    \vspace{-0.2cm}
\end{table*}

% \begin{figure*}
% \begin{center}
% \center{\includegraphics[width=\textwidth]{eval_graph.png}}
% \end{center}
%   \caption{Qualitative representation of validation losses for different evaluation tasks.}
% \label{fig:eval_graph}
% \end{figure*}

\begin{figure*}
\begin{center}
\center{\includegraphics[width=\textwidth]{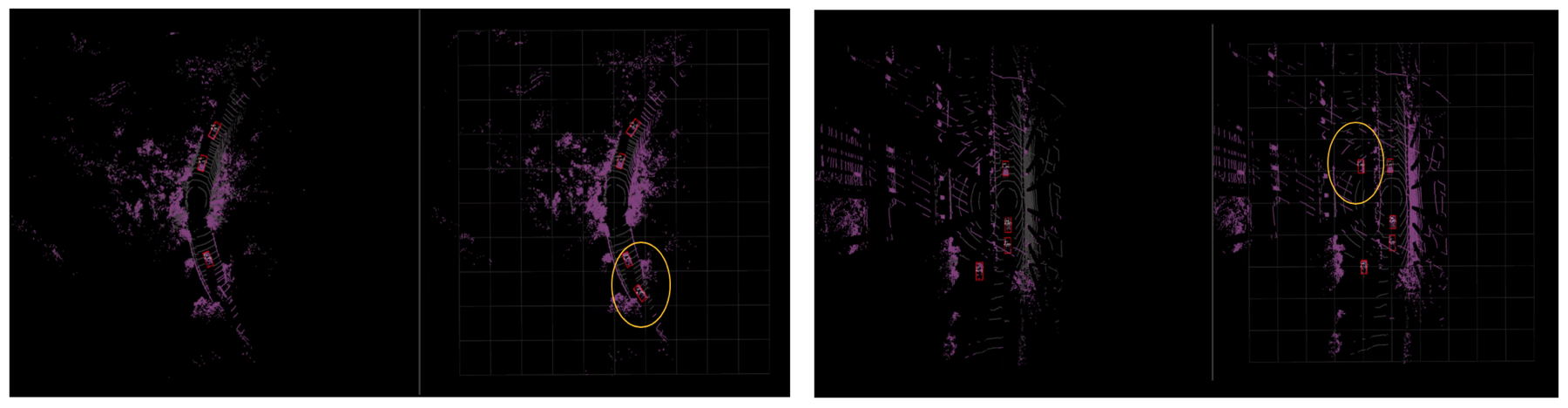}}
\end{center}
   \caption{Qualitative representation of the object detection comparing ~\cite{ku2018joint}(* lidar only)(left) with the proposed method (right). The highlighted (yellow) region shows few instances where the object detection is improved due to better ground representation.}
\label{fig:od_diff}
\end{figure*}

The training is done in an end-to-end manner, where all components including the feature extraction and ground surface estimation are defined as a part of the tensor graph. While the proposed ground segmentation algorithm does not have any trainable parameters, initializing them as a part of the graph provides significant gains in computation, and are easy to deploy in the autonomous vehicle. 

\subsection{Loss Functions}\label{ssec:loss_function}
\vspace{-0.05 in}
For both BEV (2D) and 3D object detection, we maximize the overlap between the estimated and ground truth object annotations. If the Intersection over Union (IOU) is greater than a predetermined threshold ($\tau$), then the candidate is considered as a detection (for both BEV(2D) and 3D). For regression of orientation, we use a weighted L1 loss where the regressed angle $\theta$ is converted to {\it xy}-coordinates, alongside an orientation classifier that uses cross-entropy loss for classifying the direction of the orientation.

\subsection{Implementation Details}\label{ssec:implementation_details}
\vspace{-0.05 in}
The region of interest used in the experiments along {\it y}-direction and {\it x}-direction (in BEV) are {\it 100m} and {\it 60m} respectively, around the ego vehicle. The sampling rate for generating voxel grid is set to {\it 0.1} for feature extraction and ground segmentation, and {\it 0.5} for anchor placement. Also we prune any Lidar points that are outside the {\it 100m}$\times${\it 60m} boundary, as well as we ignore points whose height lie outside of the {\it -3m to +3m} range {\it wrt.} the Lidar center.

For training, we use Adam optimizer with the initial learning rate of {\it 1e-5} with momentum update of $0.8$ for every $30,000$ steps to a total of $200,000$ steps, where throughout our experiments the batch size is 1. Also, as mentioned above, we use the {\it 3D box} representation for regressing the 3D bounding box. Though the {\it 3D Box} representation is shown as not as efficient as {\it 4CA} by~\cite{ku2018joint}, we were able to identify one of the key problems that has caused the deterioration in performance, when used 4-corners representation, the weights for {\it x, y} and {\it z} are evenly distributed and thus are the orientation as the orientation is implicitly captured via the {\it x, y, \& z}, which is not quite the case for 3D representation in which only one parameter out of 7 are dedicated for orientation. By learning the corresponding weights accordingly, we were able to ultimately obtain superior performance when using 3D box representation as shown in Tables~\ref{tab:compare_distance} \&~\ref{tab:compare_density}.

\vspace{-0.1 in}
\section{Evaluation \& Discussion}\label{sec:evaluation}
\vspace{-0.05 in}
We evaluate the performance of the proposed method for two primary problems, BEV and 3D object detection, on subsets of KITTI data categorized as {\it easy, medium} and {\it hard} which are done based on criteria detailed in section~\ref{sec:dataset}. The average cost of computing (in {\it ms}) the proposed ground surface estimation algorithm (in Section~\ref{ssec:ground_segmentation}) is {2.83ms} which is significantly lesser ($\sim$10x) than {\it 22.5ms}, the average cost of plane estimation algorithm proposed by~\cite{bogoslavskyi2017efficient}. Thus, in total, the average inference time for ~\cite{ku2018joint} is {\it 83.4ms}, whereas the average inference time for the proposed method is {\it 61.5ms}, while the computational gains predominantly come from the proposed ground surface estimation algorithm. 

For both BEV and 3D Object detection, an object candidate considered a valid detection if the Intersection Over Union (IOU) is greater than or equal to $\tau$. Table~\ref{tab:Compare} shows the comparison of performance of the proposed method against MV3D~\cite{chen2017multi}, VoxelNet~\cite{zhou2018voxelnet}, F-PointNet~\cite{qi2018frustum} and AVOD~\cite{ku2018joint}, where the proposed method is evidently the fastest, and yet the performance is comparable to others. Furthermore, we experimented with different values of $\tau = \{0.5$ \& $0.7\}$ in this paper. The comparison between different IOUs are demonstrated in Tables~\ref{tab:compare_density} \& ~\ref{tab:compare_density_05iou}. While an IOU of {\it 0.7} is commonly used, we showcased that the mis-classifications that occur in the proposed model, are considerably improving if the IOU threshold is relaxed by {\it 0.2} (which roughly translates to 20-30cm above or below in real objects scale). Thus, it is evident that the mis-classifications are not completely erroneous object detections, but occur due to the lack of sufficient overlap in prediction. Also both in Table~\ref{tab:compare_density_05iou} \& ~\ref{tab:compare_density}, we clearly demonstrate that with simply replacing the ground surface estimation module, with largely preserving most of the AVOD~\cite{ku2018joint} architecture, we are able to improve the accuracy significantly by over 15\% on {\it hard} objects.
In addition, all experiments in the paper use {\it 3D box} regression which is shown to be relatively ineffective in comparison to $4CA$~\cite{ku2018joint}. However, with optimizing the weights (of orientation parameter) in loss function, and by adding a direction classifier for classifying orientation enabled us to obtain a better performance {\it wrt.} the ~\cite{ku2018joint} baseline.

\vspace{-0.05 in}
\section{Conclusion}
\vspace{-0.05 in}
In this paper, we proposed a novel adaptive ground representation as a part of an end-to-end trainable deep learning architecture. The proposed method is computationally efficient, and is shown to be far more accurate in computing ground segmentation in comparison to its counterparts, while being $\sim$10x faster. We have successfully demonstrated the performance of the proposed architecture both qualitatively and quantitatively to be better than other two-stage Lidar object detection pipelines. We also have shown that the trained model not only works well on the dataset, but also on real data collected with a slightly different sensor configuration (Figures~\ref{fig:min_surface_perf} \&~\ref{fig:od_diff}). 

{\small
\bibliographystyle{ieee_fullname}
\bibliography{egbib}
}

\end{document}